\documentclass[10pt,conference]{IEEEtran}

\usepackage{multirow}

\usepackage{cite}

\ifCLASSINFOpdf
   \usepackage[pdftex]{graphicx}
   \graphicspath{{figs/}}
   \DeclareGraphicsExtensions{.pdf,.jpeg,.png}
\else
   \usepackage[dvips]{graphicx}
   \graphicspath{{../figs/}}
   \DeclareGraphicsExtensions{.eps}
\fi
\usepackage[normalem]{ulem}\usepackage{xcolor}

\usepackage[cmex10]{amsmath}
\usepackage{amsthm}

\usepackage{amssymb} 

\usepackage{algorithmic}
\usepackage{algorithm}
\renewcommand{\algorithmiccomment}[1]{\hfill $\triangleright$ #1}

\usepackage{array}

\ifCLASSOPTIONcompsoc
  \usepackage[caption=false,font=normalsize,labelfont=sf,textfont=sf]{subfig}
\else
  \usepackage[caption=false,font=footnotesize]{subfig}
\fi
\usepackage{url}

\newif\iffinal
\finaltrue

\iffinal
\else
\usepackage[switch]{lineno}
\fi

\begin{document}
%
\title{Efficient License Plate Recognition in Videos Using Visual Rhythm and Accumulative Line Analysis}


\iffinal

\author{
\IEEEauthorblockN{Victor Nascimento Ribeiro, Nina S. T. Hirata}
\IEEEauthorblockA{University of São Paulo - USP\\
Institute of Mathematics and Statistics\\
SP - São Paulo, Brazil}
}


%


\maketitle

\begin{abstract}
  Video-based Automatic License Plate Recognition (ALPR) involves extracting vehicle license plate text information from video captures. Traditional systems typically rely heavily on high-end computing resources and utilize multiple frames to recognize license plates, leading to increased computational overhead. In this paper, we propose two methods capable of efficiently extracting exactly one frame per vehicle and recognizing its license plate characters from this single image, thus significantly reducing computational demands. The first method uses Visual Rhythm (VR) to generate time-spatial images from videos, while the second employs Accumulative Line Analysis (ALA), a novel algorithm based on single-line video processing for real-time operation. Both methods leverage YOLO for license plate detection within the frame and a Convolutional Neural Network (CNN) for Optical Character Recognition (OCR) to extract textual information. Experiments on real videos demonstrate that the proposed methods achieve results comparable to traditional frame-by-frame approaches, with processing speeds three times faster.
\end{abstract}


\IEEEpeerreviewmaketitle

\section{Introduction}


Automatic License Plate Recognition (ALPR) systems are crucial for modern transportation infrastructure, in road traffic monitoring and law enforcement~\cite{alpr-its}. Designing these systems is challenging due to diverse license plate formats and factors such as plate tilt, blurring, and lighting variations~\cite{alpr-difficult-indian,challenges}. 

Early ALPR methods relied on image processing techniques like edge detection, morphological operations, and template matching, as surveyed by Anagnostopoulos \emph{et al.}~\cite{alpr-difficulties}. These methods often struggled with lighting variations, occlusions, and diverse plate designs. Recent advancements have incorporated deep learning models. Lin \emph{et al.}~\cite{challenges} developed an edge-AI-based ALPR system using YOLO, achieving real-time processing and high precision in various environmental conditions. This approach significantly reduced the computational overhead compared to earlier methods.

In real-world scenarios, video-based ALPR systems often rely on multiple frames rather than single images~\cite{alpr-multiple-frames} since motion blur, partial occlusion, or poor lighting may harm accurate detection from single frames. Using multiple frames allows for information aggregation, improving robustness and accuracy. Techniques like object tracking and frame selection help identify the most informative frames, ensuring effective handling of diverse and dynamic traffic scenarios. Wang \emph{et al.}~\cite{alpr-multiple-frames} introduced a large-scale video-based license plate detection and recognition system (LSV-LP), a system using object tracking and CNN-based recognition to handle motion blur and varying vehicle speeds, achieving state-of-the-art performance.  
However reliance on multiple frames makes them computationally expensive. 

We propose two methods for ALPR that explores the idea of selectively extracting and processing a single frame per vehicle in video data. The first approach is based on Visual Rhythm, a technique that generates time-spatial images from videos, and the second approach employs the Accumulative Line Analysis algorithm (ALA), a robust single-line video processing method designed for real-time operation. Both achieve efficiency focusing on monitoring a single fixed line in the frames to detect vehicles. This enables efficient license plate character recognition from a single image, reducing computational overhead.



\section{Background}
\label{sec-bg}


\subsection{Visual Rhythm}
\label{sec:VR}

Processing videos frame-by-frame to detect instants when vehicles and their license plates are visible is computationally expensive due to redundant computations and high memory usage~\cite{frame-hopper}. We can employ Visual Rhythm (VR) to address this by generating time-spatial images from videos, leveraging temporal consistency and spatial patterns across frames to create a condensed representation of the video's content over time~\cite{vr-old, vr-definition}.


Let $V$ denote an arbitrary video segment with $T$ frames, each of size $N \times M$ (width and height, respectively). Here, $f_t$ represents the frame of the video at time $t$, $t = 1, \dots, T$, and $f_t(x,y)$ denotes the pixel value at position $(x,y)$, with $0 \leq x < N$ and $0 \leq y < M$ within the frame.

Consider a scenario where a static camera captures a video from a top-view perspective of vehicles in unidirectional vertical movement within the frame. The VR method transforms the video into a two-dimensional image by stacking pixels from each frame along a predefined horizontal line at height $\lambda \in \mathbb{Z}$. 
 This forms a VR image of size $N \times T$, creating a time-spatial representation of the video, as defined in Eq.~\ref{eq} and illustrated in Fig.~\ref{fig:vr-generation}. 

\begin{equation} \label{eq}
VR(z,t) = f_t(z,\lambda), \; 0 \leq z < N
\end{equation}

The top part of Fig.~\ref{fig:vr-generation} shows five frames of a video sequence, with the predefined line superimposed in green (best viewed in electronic format). The bottom part shows the resulting time-spatial VR image (time dimension in the vertical axis) of the entire video sequence. Each time a vehicle crosses the predefined line, a mark appears in the VR image at the corresponding row of the frame index while maintaining the spatial information (horizontal axis). This allows us to easily identify the frame in which the vehicle crosses the line by observing the position of each mark.
It is important to note, however, that certain marks in the VR image may not correspond to vehicles; for instance, any object crossing the line would produce a corresponding mark.

\begin{figure}[htp]
    \centering
    \includegraphics[width=1\linewidth]{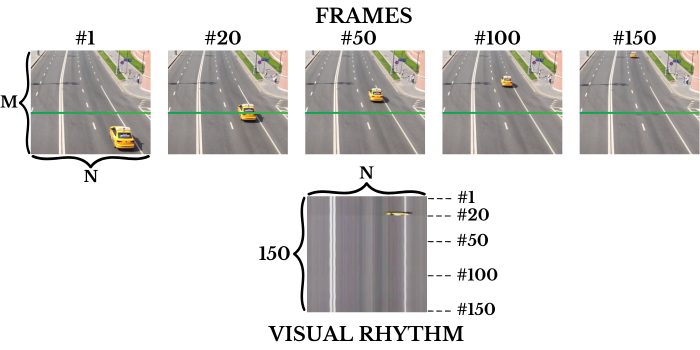}
    \caption{VR image building; example for a video sequence of 150 frames.}
    \label{fig:vr-generation}
\end{figure}

\subsection{YOLO}

YOLO (You Only Look Once)~\cite{yolo-v1} is a state-of-the-art deep learning model for object detection. It processes the image in a single pass through a neural network, predicting bounding boxes and class probabilities simultaneously~\cite{v1-v8}. YOLO's efficiency and speed make it ideal for real-time applications like video surveillance and vehicle license plate detection, where rapid processing and high accuracy are essential.

\subsection{Optical Character Recognition}

Optical Character Recognition (OCR) converts various types of documents, such as scanned paper documents or digital images, into editable and searchable text. In ALPR, OCR extracts textual information from detected license plates.

OCR systems typically involve preprocessing, character segmentation, feature extraction, and classification~\cite{ocr-overview}. Convolutional Neural Networks (CNNs) are often used for OCR in ALPR due to their effectiveness in image recognition tasks~\cite{cnn-in-ocr-systems}. CNN-based OCR models  efficiently handle challenges like varying fonts, sizes, orientations, lighting conditions, and image quality.

\section{Proposed Methods}

To circumvent the need for multiple frames to recognize license plates, we propose two distinct methods capable of selectively identifying video frames containing vehicles with visible license plates. 
Both methods are designed under the assumption that objects move unidirectionally, cross a predefined line in top-view videos, and maintain velocities within the camera's frame rate \cite{VR-plankton}.

\subsection{Visual Rhythm based method}

We integrate Visual Rhythm with YOLO, as shown in Fig.~\ref{fig:vr-method}.
In step (a), we create a Visual Rhythm image (see Section~\ref{sec:VR}) for segments of $T$ consecutive frames of an input video sequence. Assuming a vehicle $i$ intersects the predefined line in $T_i$ successive frames, the resulting VR image will contain a mark corresponding to vehicle $i$ with height equal to $T_i$ and width equal to the width of the vehicle. 

\begin{figure}[htp]
    \centering
    \includegraphics[width=1\linewidth]{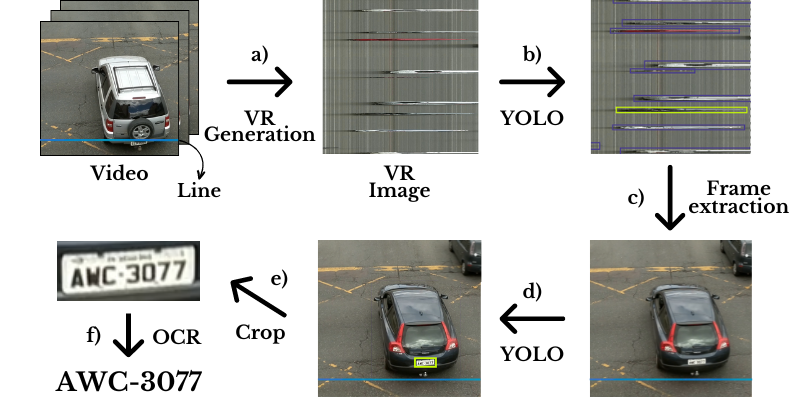}
    \caption{Data flow in the VR–based ALPR system}
    \label{fig:vr-method}
\end{figure}

In step (b), we employ YOLO to detect each of the marks within the VR image. For each detected mark, the corresponding frame from the video segment is extracted (step (c)). In the example, the extracted frame corresponds to the mark highlighted in yellow in the VR image. The $y$-coordinate of the mark's bottom corresponds to the temporal index of the frame when the vehicle has entirely crossed the line. Then, we use YOLO to detect all license plates within the extracted frame (step (d)). For each detected license plate, we verify if its bounding box's $x$-coordinates fall within the mark's $x$-coordinates. The license plate that best matches the position of the mark and is closest to the line is selected and cropped (step (e)). Finally, we employ an OCR model to extract the license plate text (step (f)).

\subsection{Accumulative Line Analysis Algorithm}

Accumulative Line Analysis Algorithm (ALA), a novel approach for ALPR, utilizes robust single-line processing based on background subtraction and logical operations. The core idea of the algorithm is to focus on the predefined horizontal line per frame of a video. By monitoring the successive patterns in the line through background subtraction, it attempts to keep track of frames in which a vehicle is crossing the line and then identify the moment right after the vehicle finishes crossing the line. Then, at this moment the current frame is extracted for license plate recognition.



\begin{algorithm}[htb]
    \caption{Accumulative Line Analysis Algorithm}
    \label{alg:realtime}
    \begin{algorithmic}[1]
        \STATE $\text{line\_or} \leftarrow \{0\}$ 
        \FOR{$\text{frame}$ \textbf{in} $\text{Video}$}       
            \STATE $\text{line} \leftarrow \text{frame}[\lambda]$ \algorithmiccomment{Get pixels of the predefined line}
            \STATE $\text{line} \leftarrow \text{BackgroundSubtractor}(\text{line})$ \label{fun:bg-sub}
            
            \STATE $\text{line\_or} \leftarrow \text{line} \lor \text{line\_or}$
            
            \STATE $\text{clusters} \leftarrow \text{getClusters} (\text{line\_or})$ \label{fun:clusters}
            \FOR{$(l, r)$ \textbf{in} $\text{clusters}$}
                \STATE $\text{line\_xor} \leftarrow \text{line}[l:r] \oplus \text{line\_or}[l:r]$
                
                \IF{$r-l < \gamma$} 
                    \STATE $\text{line\_or}[l:r] \leftarrow \{0\}$ \label{if:rem-noise} \hfill \algorithmiccomment{Delete potential noises}
                \ELSIF{$\text{sum}(\text{line\_xor}) = r-l$} 
                    \STATE $\text{ALPR}(\text{frame})$ \label{fun:alpr} \algorithmiccomment{Recognize license plate}
                    \STATE $\text{line\_or}[l:r] \leftarrow \{0\}$
                \ENDIF
            \ENDFOR
        \ENDFOR
    \end{algorithmic}
\end{algorithm}

As outlined in Algorithm~\ref{alg:realtime}, in line 3 the \textit{BackgroundSubtractor(line)} produces a foreground-masked binarized representation of '\textit{line}'. Any background subtraction algorithm can be used here. The algorithm accumulates the presence of pixels (indicating parts of the vehicle) on the predefined line across frames using a logical OR operation. This accumulated line (\textit{line\_or}) helps in tracking the continuous movement of the vehicle. In line 6, invoking the \textit{getClusters(line\_or)} function returns an array of indices denoting the start ($l$) and end ($r$) of each cluster within '\textit{line\_or}'. By identifying clusters of pixels in the accumulated line, the algorithm determines segments where a vehicle has potentially crossed the line.

The XOR operation is critical for determining when the vehicle has entirely crossed the line. The XOR operation between the current line (\textit{line}) and the accumulated line (\textit{line\_or}) within a cluster checks for changes in the vehicle’s presence across frames. If the XOR sum result is equal to the size of the cluster $(r-l)$, it means that the current frame's line pixels completely differ from the accumulated line pixels within that cluster, indicating that the vehicle has fully crossed the line. In line 12, the \textit{ALPR(frame)} function extracts textual information of the license plate from '\textit{frame}', employing the same procedure as steps (d) onward in the VR method depicted in Figure~\ref{fig:vr-method}. To ensure robust detection, the algorithm discards clusters smaller than a predefined threshold $\gamma$ to eliminate noise.

This algorithm is particularly effective for real-time applications due to its simplicity and efficiency, reducing the computational load by focusing on a single line per frame rather than the entire frame.

\section{Results and discussion}

The experiments were conducted on a system running Python and Ultralytics. The hardware configuration included an Intel Xeon Silver 4216 2.10GHz CPU and an NVIDIA RTX A5000 GPU. The source code is publicly available at \url{https://github.com/victor-nasc/Vehicle-Licence-Plate-Recognition}.

\subsection{Preparation}

In this study, we used the YOLOv8-small model, fine-tuning it on task-specific datasets to enhance detection performance. For mark detection within the VR method, we generated 300 VR images from the four videos in the dataset described in \cite{eu-mesmo}. For license plate detection, we fine-tuned the model on a publicly available dataset of over 10,000 images \cite{lp-dataset}.

In the Accumulative Line Analysis (ALA) approach, we used the Mixture of Gaussians 2 (MOG2) method for background subtraction with a history parameter set to 1, ensuring the model responds quickly to immediate scene changes.

For Optical Character Recognition (OCR), we utilized the CNN-OCR model~\cite{cnn-ocr}, an architecture based on CR-NET and fine-tuned for the Brazilian license plate layout (three letters followed by four digits). The model includes the first eleven layers of YOLO and adds four convolutional layers to enhance non-linearity, effectively predicting 35 character classes~\cite{vehicle-rear}.

We evaluated our methods using the Vehicle-Rear dataset \cite{vehicle-rear}, which consists of 5 high-resolution videos meeting the constraints of both methods, with clearly visible vehicle license plates. These videos capture a busy cross-intersection. For testing purposes, we manually selected only from footage of Camera 1 the segments where vehicles move vertically within the frame. The test set contains a total of 1,512 vehicles with visible license plate, providing a robust evaluation. Some example images are available on GitHub.





\subsection{Results}

For the experiments, we set $\gamma = 100$ due to the video resolution and to build the VR images we used segments of $T = 900$ frames (30 seconds), an optimal size for YOLO processing. The line was positioned at a height of $\lambda = 1000$ pixels on the original videos, ensuring coverage of the entire vehicle path. Table \ref{tab:frame-extraction} compares Precision (P), Recall (R), and F-score (F) for vehicle frame extraction.

\begin{table}[htp]
\renewcommand{\arraystretch}{1.3}
\caption{Vehicle Frame Extraction Results}
\label{tab:frame-extraction}
\centering
\begin{tabular}{|c|ccc|ccc|}
    \hline
    \multirow{2}{*}{\textbf{Video}} & \multicolumn{3}{c|}{\textbf{Visual Rhythm}} & \multicolumn{3}{c|}{\textbf{ALA}} \\
    \cline{2-7}
    & \textbf{P} & \textbf{R} & \textbf{F} & \textbf{P} & \textbf{R} & \textbf{F} \\
    \hline
    01 & 94.9\% & 97.4\% & 96.1\% & 84.1\% & 94.2\% & 88.8\% \\
    02 & 95.6\% & 96.5\% & 96.0\% & 86.0\% & 86.4\% & 86.2\% \\
    03 & 99.4\% & 99.0\% & 99.1\% & 99.2\% & 98.4\% & 98.8\% \\
    04 & 98.8\% & 98.1\% & 98.4\% & 97.8\% & 97.1\% & 97.4\% \\
    05 & 99.2\% & 98.9\% & 99.0\% & 99.3\% & 96.0\% & 97.6\% \\
    \hline
\end{tabular}
\end{table}

The results indicate that the VR approach consistently outperforms the ALA algorithm across all videos. Specifically, Videos 3, 4, and 5 show minimal variation in lighting conditions, resulting in better performance for both methods. 

In contrast, Videos 1 and 2 exhibit significant lighting variations and shadows due to the sun's angle, which negatively affects both methods, particularly the ALA algorithm that relies heavily on background subtraction, and its performance is impacted more by shadows and lighting changes. As a result, ALA shows lower Precision and Recall in these videos, with F-scores around 88.8\% and 86.2\%, respectively. The VR method, however, maintains higher Precision and Recall, demonstrating its robustness to such variations, with F-scores of 96.1\% and 96.0\% for Videos 1 and 2, respectively.

Overall, the Visual Rhythm method offers a more reliable and consistent approach for vehicle frame extraction in ALPR systems, particularly under varying lighting conditions.

We compared the efficiency of our approaches with Roboflow's frame-by-frame vehicle counting system~\cite{bytetrack}. This system uses the ByteTrack tracking algorithm for real-time object counting and has been adapted for ALPR. Both systems were tested under identical environments using the same license plate detection model weights. Table \ref{tab:ocr-results} shows the mean OCR accuracy comparison for all three methods, with accuracy calculated with respect to recognition of individual characters.

\begin{table}[htp]
\renewcommand{\arraystretch}{1.3}
    \caption{License Plate Mean OCR Results}
    \label{tab:ocr-results}
    \centering
    \begin{tabular}{|c|c|c|}
        \hline
        \textbf{System} & \textbf{Frame rate} & \textbf{Accuracy}\\
        \hline
        Bytetrack~\cite{bytetrack} & 24 FPS & 88.72\%\\
        Visual Rhythm             & 63 FPS & 88.66\%\\
        ALA                       & 74 FPS & 80.78\%\\
        \hline
    \end{tabular}
\end{table}

Both the Visual Rhythm and ALA methods demonstrate significant speed improvements compared to traditional frame-by-frame approaches, with frame rates of 63 FPS and 74 FPS, respectively. Despite the higher speed, Visual Rhythm achieves similar OCR accuracy (88.66\%) to the ByteTrack method (88.72\%), while ALA, although slightly faster, it has a much lower accuracy (80.78\%).

It's important to note that if one of the early steps fails and it becomes impossible to extract the license plate, we consider the approach to have misread every character in the license plate for the accuracy calculation.

\section{Conclusion}

In this paper, we proposed two time-efficient methods for Automatic License Plate Recognition (ALPR) in videos: Visual Rhythm (VR) and Accumulative Line Analysis (ALA). These methods extract a single frame per vehicle and accurately recognize its license plate characters, minimizing computational overhead.

Comparative analysis with the ByteTrack frame-by-frame vehicle counting system highlighted the efficiency of our methods. Both the VR and ALA methods achieved higher frame rates, with VR maintaining comparable OCR accuracy to ByteTrack. This indicates that our proposed methods not only reduce computational complexity but also deliver reliable performance in real-world video-based ALPR applications.

Overall, the Visual Rhythm method provides a more reliable and consistent approach for vehicle frame extraction and license plate recognition than ALA, especially under challenging environmental conditions. Furthermore, VR maintains comparable OCR results to traditional multi-frame approaches. However, there is ample room for improvement in the ALA algorithm to enhance its robustness and performance under varying lighting conditions and other challenging scenarios.

\section*{Acknowledgment}

MCTI (\textit{Ministério da Ciência, Tecnologia e Inovações, Brazil}), law 8.248, PPI-Softex - TIC 13 - 01245.010222/2022-44, \textit{Fundação de Apoio à Universidade de São Paulo} (FUSP), and FAPESP (grant 2022/15304-4).



\bibliographystyle{IEEEtran}
\bibliography{sibgrapi2024}
%
%


\end{document}